\documentclass[10pt,twocolumn,letterpaper]{article}

\usepackage{cvpr}              

\usepackage{balance}

\definecolor{cvprblue}{rgb}{0.21,0.49,0.74}
\usepackage[pagebackref,breaklinks,colorlinks,allcolors=cvprblue]{hyperref}
\usepackage{multirow}

\makeatletter
\renewcommand\section{\@startsection{section}{1}{\z@}
  {-6pt plus -2pt minus -2pt}{4pt plus 1pt minus 1pt}
  {\normalfont\large\bfseries}}
\renewcommand\subsection{\@startsection{subsection}{2}{\z@}
  {-5pt plus -2pt minus -2pt}{3pt plus 1pt minus 1pt}
  {\normalfont\normalsize\bfseries}}
\makeatother

\def\paperID{6} 
\def\confName{CVPR}
\def\confYear{2026}

\title{Diverse by Design: Architectural Constraints for Prototype-Based Interpretability}

\author{Xinmiao Lin\thanks{Work done prior to joining Amazon.}\\
Rochester Institute of Technology\\
{\tt\small xl3439@rit.edu}
\and
Matthew Wright\\
Rochester Institute of Technology\\
{\tt\small matthew.wright@rit.edu}
}

\begin{document}
\maketitle


\begin{abstract}
Prototype-based neural networks provide inherent interpretability through case-based reasoning, yet suffer from critical limitations: prototypes converge to redundant features, fail to capture diverse semantic parts, and lack quantitative interpretability assessment. We propose Diversity-Aware Prototype Learning (DAPL), which enforces prototype diversity through architectural constraints rather than explicit regularization. Our approach leverages multi-head self-attention with strict one-to-one attention-to-prototype mapping, ensuring each prototype specializes in distinct visual features. We further introduce foreground-aware training to focus prototypes on semantically meaningful regions and develop comprehensive evaluation metrics (Coverage and Diversity) for quantitative interpretability assessment. Experiments on CUB-200-2011 demonstrate substantial improvements: DAPL with foreground-aware training achieves 81.69\% accuracy with 0.596 Coverage and 0.427 Diversity, providing the best overall balance across all evaluated prototype-based methods. Code is available at \url{https://github.com/xinmiaolin/DAPL}.

\end{abstract}

\section{Introduction}
\label{sec:intro}

Deep learning has achieved remarkable success in computer vision, but its black-box nature limits deployment in high-stakes domains requiring transparency. Prototype-based neural networks~\cite{this-that,protopshare,pixel-grounded} offer an inherently interpretable alternative: they learn prototypical visual patterns and classify by measuring similarity to learned prototypes, providing case-based reasoning that mirrors human cognition.

\begin{figure}[t]
\centering
\includegraphics[width=1.0\linewidth]{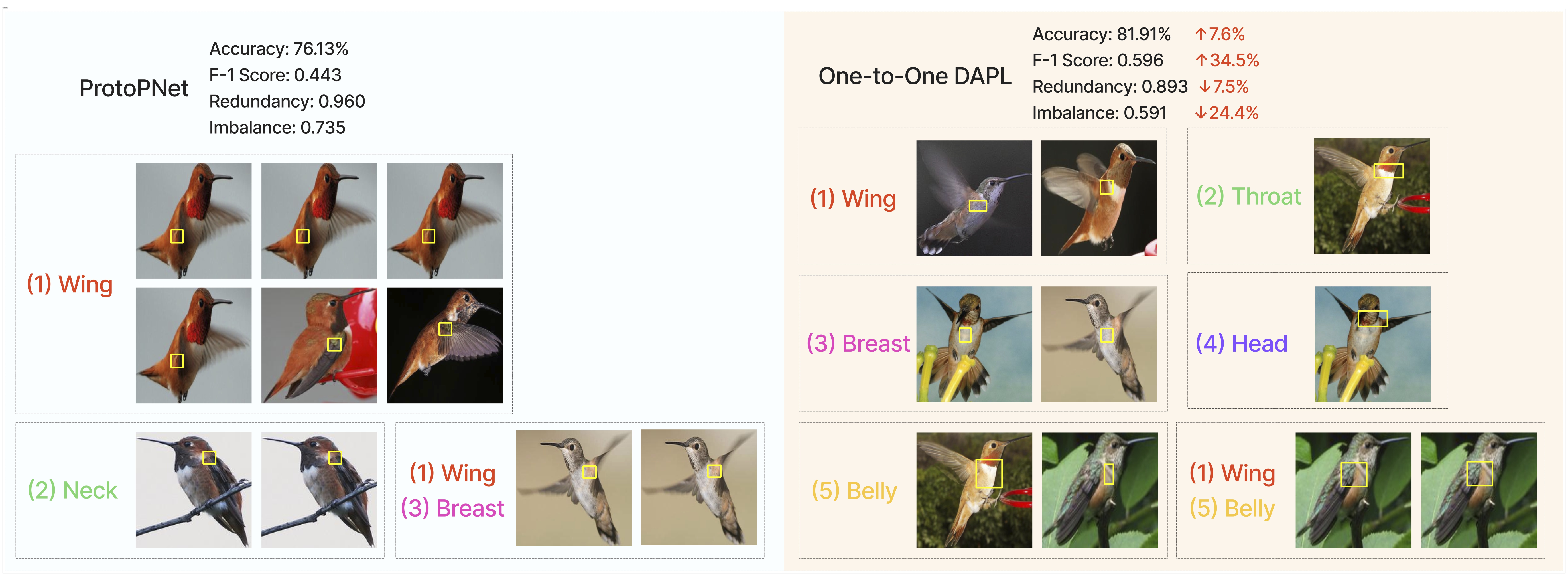}
\caption{\textbf{Motivation: Prototype redundancy in existing methods.} ProtoPNet learns multiple redundant prototypes (left) focusing on similar wing regions, while DAPL (right) achieves diverse coverage of anatomical features.}
\label{fig:motivation}
\end{figure}

Existing methods, however, suffer from prototype redundancy. Optimization dynamics favor learning multiple prototypes capturing similar discriminative features, because standard objectives lack explicit diversity mechanisms. As shown in Figure~\ref{fig:motivation}, ProtoPNet learns 6 of 10 prototypes on nearly identical wing patterns while missing other anatomical features. Moreover, prototypes often activate on background regions rather than meaningful object parts.

We address these limitations with \emph{Diversity-Aware Prototype Learning (DAPL)}, which enforces prototype diversity via architectural constraints rather than explicit regularization. Our key insight is that multi-head self-attention naturally learns diverse spatial relationships---each head specializes in different visual patterns without explicit supervision. By enforcing strict one-to-one mapping between attention heads and prototypes, we architecturally guarantee this specialization transfers to the prototype level, preventing the redundancy that emerges from unconstrained optimization. We complement this structural constraint with contrastive regularization that explicitly maximizes inter-head prototype distinctiveness, ensuring each attention head learns truly unique visual patterns.

Beyond the architecture, we introduce \emph{foreground-aware training} that penalizes prototype activations on background regions. While background context can provide useful cues for classification, over-reliance on background features reduces model transparency. Our foreground loss encourages the model to weight object-relevant features more heavily while still allowing contextual information to contribute.

\begin{figure*}[t]
\centering
\includegraphics[width=0.95\linewidth]{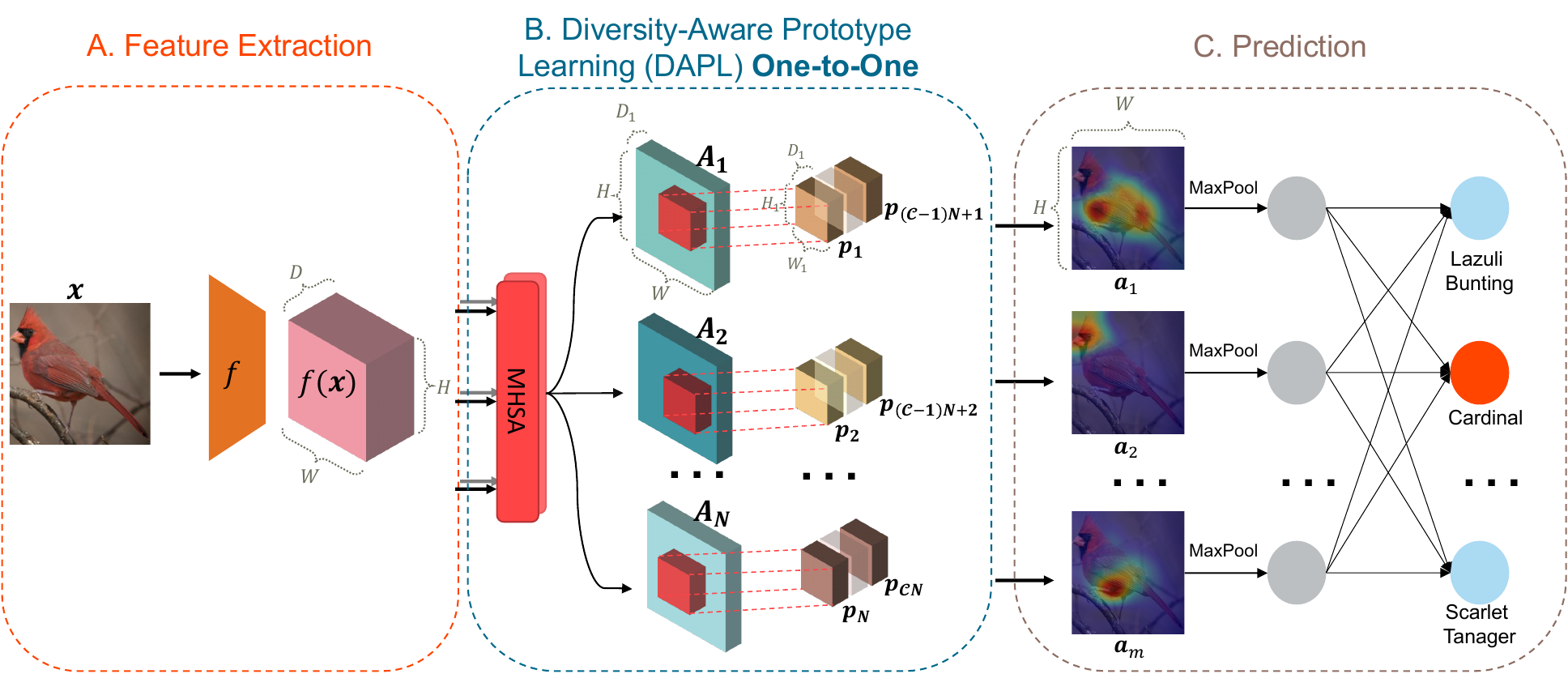}
\caption{\textbf{DAPL Architecture.} (A) CNN backbone extracts features. (B) Multi-head self-attention produces $N$ diverse outputs. (C)~One-to-one mapping assigns each attention head to specific prototypes. (D) Similarity computation drives classification.}
\label{fig:architecture}
\end{figure*}

To assess these ideas, we propose two new metrics: 
\emph{Coverage} and \emph{Diversity}. Coverage quantifies how well prototypes align with semantic object parts from ground-truth part annotations, measuring whether the learned representations capture meaningful visual concepts. Meanwhile, Diversity assesses whether prototypes are spatially distinct and evenly distributed across different parts.

Our contributions are:
\begin{itemize}
\item A novel architecture combining multi-head self-attention with one-to-one attention-to-prototype mapping and contrastive regularization to enforce prototype diversity.
\item Foreground-aware training that guides prototypes toward semantically meaningful object regions.
\item Comprehensive evaluation metrics (Coverage and Diversity) for quantitative assessment of interpretability.
\end{itemize}

Extensive experiments on CUB-200-2011 (with additional results on FaceForensics++ in the Appendix) demonstrate substantial improvements: DAPL with foreground-aware training achieves 81.69\% accuracy with 0.596 Coverage and 0.427 Diversity, providing the best overall balance across all evaluated prototype-based methods.

\section{Related Work}

\textbf{Prototype-Based Neural Networks.} ProtoPNet~\cite{this-that} pioneered case-based reasoning via similarity-based classification. Subsequent works explore cross-class sharing~\cite{protopshare}, pixel-level grounding~\cite{pixel-grounded}, deformable prototypes~\cite{deformable-protopnet}, differentiable assignment~\cite{protopool}, attention mechanisms~\cite{protoattend}, improved visualization~\cite{this-looks-like-those}, patch-based grounding~\cite{pip-net}, and domain-specific extensions~\cite{protomil,tesnet,st-protopnet,protoargnet,protosolo,protovae,lucidppn}. However, none explicitly address prototype redundancy or provide quantitative diversity metrics.

\textbf{Attention Mechanisms for Interpretability.} Multi-head attention~\cite{attention_need,vit} naturally learns diverse feature representations. Recent advances include attention flow via Markov chains~\cite{attention-chains} and gradient-based ViT explainability~\cite{bousselham2025legrad}. Prototype-based ViTs~\cite{proto-vit,vit-net,protopformer,concepttransformer} show promise, but we focus on CNN backbones for compatibility with established baselines, leveraging attention to enforce diversity via one-to-one attention-to-prototype mapping.


\textbf{Interpretability Evaluation.} Existing metrics address stability~\cite{con_stab_metrics}, consistency and purity~\cite{pip-net}, sparsity~\cite{protopnext}, and prototype quality~\cite{evaluation-protonet}, while broader approaches rely on human studies~\cite{mohseni2021multidisciplinary}, faithfulness scores~\cite{alvarez2018robustness}, deletion metrics~\cite{petsiuk2018rise,lin2021gradient}, perturbation analysis~\cite{samek2017evaluating}, or transferable neural pathways~\cite{lin2023modelexplanationstransferableneural}. None directly assess prototype redundancy and semantic coverage. We introduce metrics for diversity and part alignment using ground-truth annotations.

\section{Method}

\subsection{Diversity-Aware Prototype Learning}

Figure~\ref{fig:architecture} illustrates our approach. Given convolutional features $f(\mathbf{x}) \in \mathbb{R}^{H \times W \times D}$, we apply multi-head self-attention with $N$ heads to produce diverse spatial representations. Each attention head is assigned to specific prototypes through strict one-to-one correspondence, ensuring architectural diversity.

Multi-head attention computes $Q = W_q f(\mathbf{x})$, $K = W_k f(\mathbf{x})$, $V = W_v f(\mathbf{x})$, then:
\begin{equation}
\mathbf{A} = \text{softmax}\left( \frac{Q K^T}{\sqrt{D_1}} \right) V \in \mathbb{R}^{N \times D_1 \times H \times W}
\end{equation}
where $D_1 = D/N$. Each $\mathbf{A}_h$ captures different spatial relationships.

For one-to-one prototype assignment, each class $c \in \{1, ..., \mathcal{C}\}$ has $N$ prototypes, one per attention head. Attention head $h$ is exclusively compared with prototypes $\{\mathbf{p}_j | j = (k-1) \times N + h, k \in \{1, ..., \mathcal{C}\}\}$. For example, with $\mathcal{C}=200$ classes and $N=10$ heads, we have 2,000 prototypes: the first 10 belong to class 1 and pair with heads 1--10 respectively, the next 10 belong to class 2, and so on. Similarity is computed via:
\begin{equation}
s_{h,j}(\mathbf{x}) = \max_{h,w} \log \left( \frac{\|\mathbf{z}_{h,w} - \mathbf{p}_j\|_2^2 + 1}{\|\mathbf{z}_{h,w} - \mathbf{p}_j\|_2^2 + \epsilon} \right)
\end{equation}

We add contrastive regularization to explicitly enforce distinctiveness:
\begin{equation}
\mathcal{L}_{\text{contrast}} = \sum_{h=1}^N \sum_{\mathbf{p}_i \in \mathbf{P}_h} \sum_{\mathbf{p}_k \in \mathbf{P}_{\bar{h}}} \max(0, m - s_{h,i} + s_{h,k})
\end{equation}
where $m$ is a margin encouraging separation between assigned and unassigned prototypes.

The similarity scores produce class logits through a fully-connected layer:
\begin{equation}
\text{logit}_c(\mathbf{x}) = \sum_{h=1}^N w_{c,h} \cdot s_{h,j_c^h}(\mathbf{x})
\end{equation}
where $w_{c,h}$ are learnable weights and $j_c^h$ denotes the prototype index for class $c$ and head $h$. Each classification decision thus aggregates evidence from diverse attention heads.

\subsection{Foreground-Aware Training}

To prevent background activation, we introduce foreground-aware training. Given foreground mask $\mathbf{m} \in \{0,1\}^{H \times W}$, we define background bias:
\begin{equation}
B(\mathbf{p}_j, \mathbf{x}, \mathbf{m}) = \frac{\sum_{h,w} (1-\mathbf{m}[h,w]) \cdot \mathbf{a}_{h,j}[h,w]}{\sum_{h,w} \mathbf{a}_{h,j}[h,w]}
\end{equation}

The foreground loss penalizes excessive background activation:
\begin{equation}
\mathcal{L}_{\text{fg}} = \frac{1}{|\mathcal{D}|} \sum_{\mathbf{x} \in \mathcal{D}} \frac{1}{m} \sum_{j=1}^m \max \left(0, B(\mathbf{p}_j, \mathbf{x}, \mathbf{m}) - \tau_{\text{bg}} \right)
\end{equation}

Total loss: $\mathcal{L}_{\text{total}} = \mathcal{L}_{\text{ProtoPNet}} + \lambda_{\text{contrast}} \mathcal{L}_{\text{contrast}} + \lambda_{\text{fg}} \mathcal{L}_{\text{fg}}$, where $\mathcal{L}_{\text{ProtoPNet}}$ is the standard ProtoPNet objective combining classification, cluster, and separation losses~\cite{this-that}.

\subsection{Evaluation Metrics}

We propose Coverage and Diversity metrics using ground-truth part annotations. We first compute an overlap score between each prototype's activation region $A_j$ and part mask $M_k$: $\text{overlap}(\mathbf{p}_j, \text{part}_k) = |A_j \cap M_k|/|M_k|$, normalized by part size to ensure small parts (e.g., beak) are not underweighted. A prototype is assigned to a part when overlap exceeds threshold $\tau$.

\textbf{Coverage} measures whether prototypes capture semantically meaningful parts via precision (fraction of prototypes assigned to $\geq 1$ part) and recall (fraction of parts captured by $\geq 1$ prototype):
\begin{equation}
\text{Coverage} = \frac{2 \cdot \text{Precision} \cdot \text{Recall}}{\text{Precision} + \text{Recall}}
\end{equation}
High Coverage indicates that prototypes align with most of the meaningful semantic features rather than missing features or capturing background information.

\textbf{Diversity} combines two components: Redundancy, which quantifies spatial overlap (IoU) between prototypes assigned to the same part—note that same-part assignment does not imply redundancy, as two ``wing'' prototypes may capture distinct subregions (e.g., wing tip vs.\ wing bar), and high IoU indicates they activate on the same pixels rather than complementary features—and Imbalance, which measures uneven distribution of prototypes across parts using the Gini coefficient. Imbalance captures the more critical failure mode where most prototypes concentrate on a few parts while others remain uncovered:
\begin{equation}
\text{Diversity} = 1 - \frac{\text{Redundancy} + \text{Imbalance}}{2}
\end{equation}
High Diversity indicates that prototypes are spatially distinct with balanced part coverage, rather than multiple prototypes redundantly focusing on the same features.

\section{Experiments}

\subsection{Experimental Setup}

\textbf{Dataset:} We evaluate on CUB-200-2011~\cite{cub-200} (11,788 bird images, 200 species, 15 part annotations). We use DenseNet-161 backbone with 2,000 prototypes (10/class), $N=10$ attention heads, and feature dimension $D_1=16$. Assignment threshold $\tau=0.2$.

\textbf{Baselines:} ProtoPNet~\cite{this-that}, ProtoPShare~\cite{protopshare}, and PixPNet~\cite{pixel-grounded}.

\subsection{Main Results}

\begin{figure*}[t]
\centering
\includegraphics[width=0.33\linewidth]{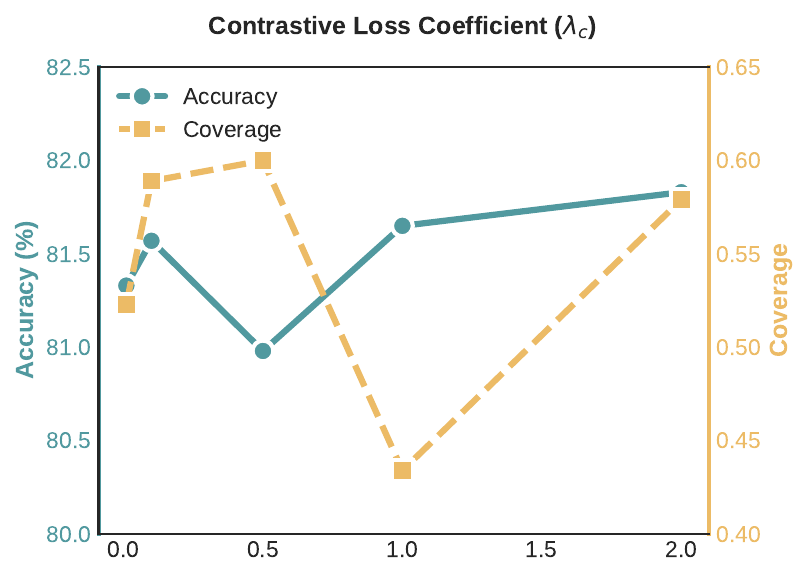}\hfill
\includegraphics[width=0.33\linewidth]{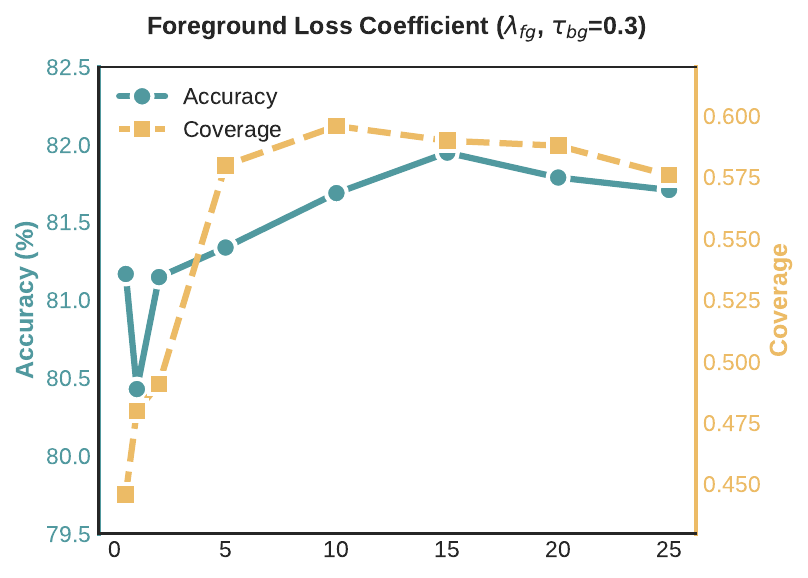}\hfill
\includegraphics[width=0.33\linewidth]{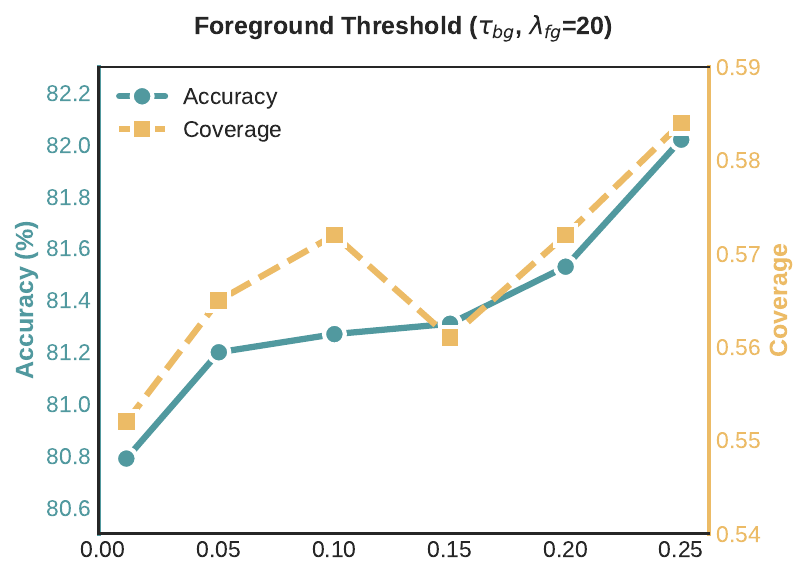}
\caption{\textbf{Ablation studies on CUB-200-2011.} (Left) Contrastive loss coefficient $\lambda_c$: moderate values achieve best Coverage. (Middle) Foreground loss coefficient $\lambda_{fg}$ (at $\tau_{bg}=0.3$): accuracy and Coverage improve with stronger guidance. (Right) Foreground threshold $\tau_{bg}$ (at $\lambda_{fg}=20$): moderate thresholds improve both accuracy and Coverage.}
\label{fig:ablation_studies}
\end{figure*}

\begin{table}[t]
\centering
\caption{Performance on CUB-200-2011 (DenseNet161).}
\label{tab:main_cub}
\begin{tabular}{lccc}
\toprule
\textbf{Method} & \textbf{Accuracy} & \textbf{Coverage} & \textbf{Diversity} \\
& (\%) & $\uparrow$ & $\uparrow$ \\
\midrule
ProtoPNet & 76.13 & 0.443 & 0.152 \\
ProtoPShare & 74.40 & 0.594 & 0.184 \\
PixPNet & 76.68 & 0.250 & \textbf{0.450} \\
\midrule
DAPL ($D_1$=32) & 80.96 & 0.514 & 0.221 \\
DAPL ($D_1$=16) & 81.57 & 0.589 & 0.277 \\
DAPL ($D_1$=8) & 80.20 & 0.580 & 0.302 \\
\midrule
DAPL ($D_1$=16) + FG & \textbf{81.69} & \textbf{0.596} & 0.427 \\
\bottomrule
\end{tabular}
\end{table}

\begin{figure}[t]
\centering
\includegraphics[width=\linewidth, trim={5.5cm 5cm 14cm 6cm}, clip]{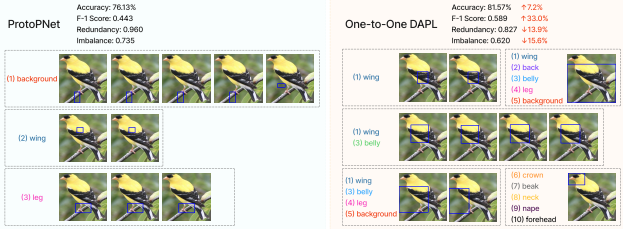}
\caption{\textbf{Prototype activation comparison.} ProtoPNet (left) shows severe redundancy with 5 background and 2 wing prototypes. DAPL (right) achieves diverse coverage across 10 distinct anatomical parts. Class: American Goldfinch.}
\label{fig:cub_comparison}
\end{figure}

Table~\ref{tab:main_cub} shows DAPL achieves \textbf{81.57\% accuracy} with \textbf{0.589 Coverage}. While PixPNet achieves highest Diversity (0.450) through pixel-level grounding, it suffers from poor Coverage (0.250). ProtoPShare achieves better Coverage (0.594) but reduced accuracy (74.40\%). DAPL with foreground-aware training achieves the best overall balance: 81.69\% accuracy, 0.596 Coverage, and 0.427 Diversity. The optimal $D_1=16$ balances fine-grained details with sufficient context; smaller dimensions ($D_1=8$) enforce Diversity but lose accuracy, while larger ones ($D_1=32$) reduce Diversity and Coverage.

Figure~\ref{fig:cub_comparison} illustrates the improvement: ProtoPNet exhibits severe redundancy with 5 background prototypes and 3 on identical wing patterns, missing critical features. DAPL achieves balanced coverage across 10 distinct anatomical regions (wing, back, belly, leg, crown, beak, neck, nape, forehead). This diversity stems from multi-head attention providing a natural architectural inductive bias---each head specializes in different spatial relationships---while our one-to-one mapping enforces this at the prototype level. Contrastive regularization further pushes prototypes toward distinct representations.

\begin{figure}[t]
\centering
\includegraphics[width=1.0\linewidth, trim={1.6cm 2cm 1.6cm 4.2cm}, clip]{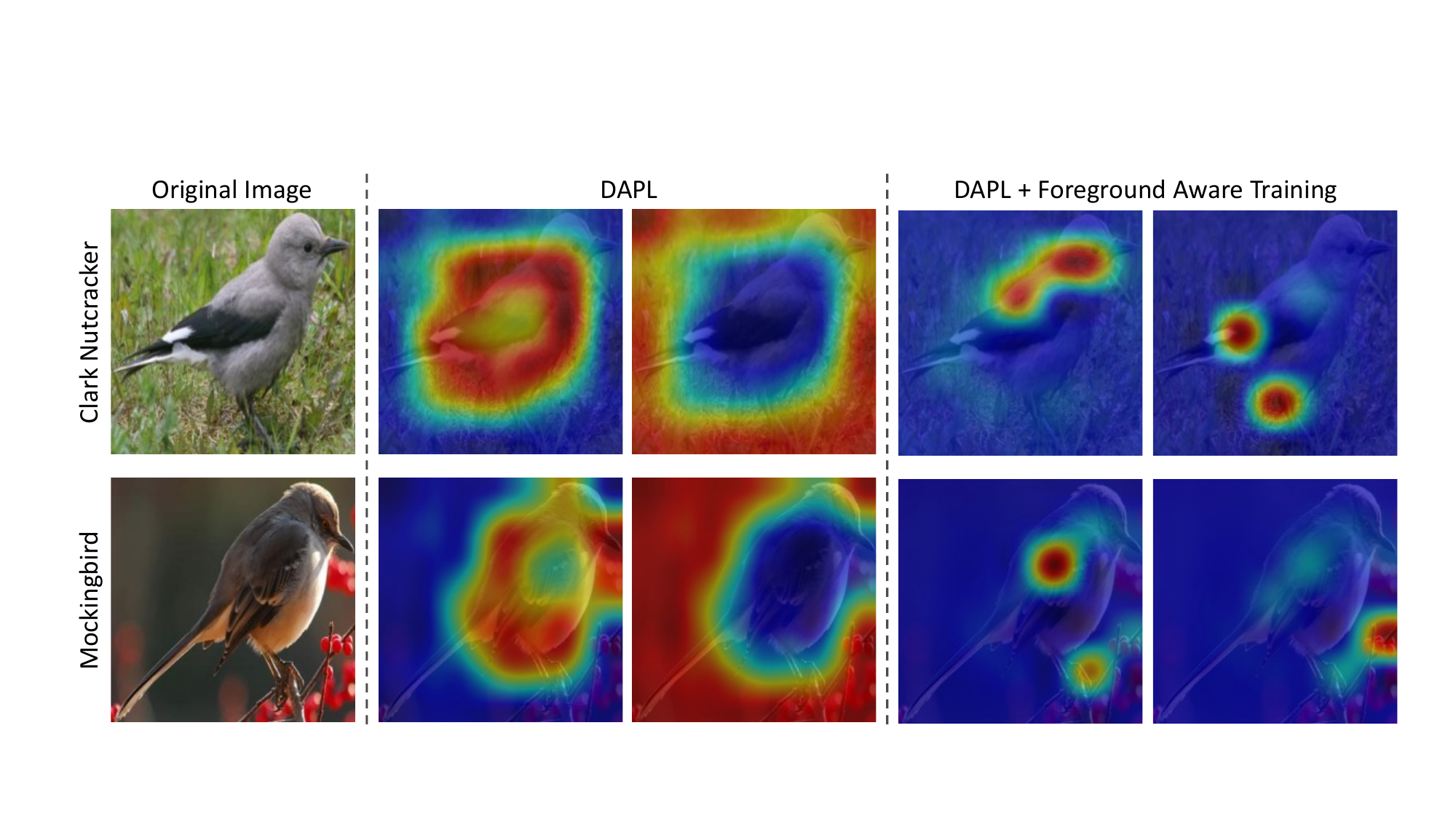}
\caption{\textbf{Foreground-aware training effect.} \textit{Top:} DAPL prototypes exhibit background bias. \textit{Bottom:} DAPL + FG focuses on meaningful bird parts.}
\label{fig:foreground_global}
\end{figure}

Figure~\ref{fig:foreground_global} shows the effect of foreground-aware training: standard DAPL activations extend diffusely into the background, while foreground training localizes them tightly on specific bird parts, explaining the Coverage improvements in Table~\ref{tab:main_cub}.

\subsection{Ablation Studies}

We analyze key hyperparameters on CUB-200-2011 (Figure~\ref{fig:ablation_studies}). Contrastive coefficient $\lambda_c$ controls the balance between diversity enforcement and classification: moderate values achieve the best Coverage (0.600 at $\lambda_c=0.5$) and accuracy (80.98\%), while very low values provide insufficient diversity and very high values over-constrain the model, suggesting that the architectural constraint should be complemented with moderate---not excessive---explicit regularization.

Foreground coefficient $\lambda_{fg}$ shows that increasing foreground guidance steadily improves both accuracy (from 81.17\% to 81.95\%) and Coverage (from 0.446 to 0.596), confirming that directing prototypes toward object features benefits both classification and interpretability. Performance plateaus around $\lambda_{fg}=10.0$-$15.0$.

Background threshold $\tau_{bg}$ reveals an important insight: some background context is beneficial. Allowing moderate background activation ($\tau_{bg}=0.20$-$0.25$) yields the best accuracy (82.02\%) and Coverage (0.584), as prototypes can capture contextual information at object boundaries. Overly strict thresholds ($\tau_{bg} < 0.10$) hurt performance by excluding valid boundary features.


\section{Conclusion}


We presented DAPL, addressing prototype redundancy through multi-head self-attention with one-to-one prototype assignment and foreground-aware training. DAPL achieves the best overall balance across accuracy (81.69\%), Coverage (0.596), and Diversity (0.427) on CUB-200-2011, demonstrating that architectural inductive biases can effectively enforce diversity in learned prototype representations. Our evaluation framework provides quantitative metrics for systematic interpretability assessment, paving the way for more trustworthy prototype-based models.

\section*{Acknowledgments}
This material is based upon work supported in part by the National Science Foundation under Award No.~2429835 and by the Omidyar Network Foundation. The authors acknowledge Research Computing~\cite{rit-rc} at the Rochester Institute of Technology for providing computational resources and support that have contributed to the research results reported in this publication.

\balance
{
    \small
    \bibliographystyle{ieeenat_fullname}
    \bibliography{workshop_submission/references}
}

\clearpage
\appendix

\section{Detailed Method Formulations}

\subsection{Multi-Head Self-Attention Details}

Given convolutional features $f(\mathbf{x}) \in \mathbb{R}^{H \times W \times D}$, we reshape them into $HW$ spatial tokens and compute queries, keys, and values:

\begin{equation}
Q = W_q f(\mathbf{x}), \quad K = W_k f(\mathbf{x}), \quad V = W_v f(\mathbf{x}) \in \mathbb{R}^{HW \times D}
\end{equation}

where $W_q, W_k, W_v \in \mathbb{R}^{D \times D}$ are learnable projection matrices. We then reshape $Q, K, V$ to separate $N$ attention heads: $\mathbb{R}^{N \times HW \times D_1}$ where $D_1 = D/N$. The attention output is:

\begin{equation}
\mathbf{A} = \text{softmax}\left( \frac{Q K^T}{\sqrt{D_1}} \right) V \in \mathbb{R}^{N \times HW \times D_1}
\end{equation}

Finally, we reshape $\mathbf{A}$ back to spatial dimensions $\mathbb{R}^{N \times D_1 \times H \times W}$. Each attention head $\mathbf{A}_i$ captures different spatial relationships, providing natural diversity.

\subsection{Similarity Computation}

For each attention head output $\mathbf{A}_h \in \mathbb{R}^{D_1 \times H \times W}$ and its assigned prototypes, we compute the activation map at each spatial location. Following the similarity computation from ProtoPNet~\cite{this-that}, we measure the distance between a patch $\mathbf{z}_{h,w} \in \mathbb{R}^{H_1 \times W_1 \times D_1}$ from the attention output at spatial location $(h,w)$ and prototype $\mathbf{p}_j$:

\begin{equation}
\mathbf{a}_{h,j}[h,w] = \log \left( \frac{\|\mathbf{z}_{h,w} - \mathbf{p}_j\|_2^2 + 1}{\|\mathbf{z}_{h,w} - \mathbf{p}_j\|_2^2 + \epsilon} \right)
\end{equation}

where $\epsilon$ is a small constant for numerical stability. The similarity score is then obtained by max-pooling over spatial locations:

\begin{equation}
s_{h,j}(\mathbf{x}) = \max_{h,w} \mathbf{a}_{h,j}[h,w]
\end{equation}

The max-pooling operation identifies the most similar patch, producing a single similarity score $s_{h,j}$ per prototype.

\subsection{One-to-One Prototype Assignment Details}

For each class $c \in \{1, ..., \mathcal{C}\}$, we allocate $N$ prototypes, one per attention head. Each prototype is exclusively compared with its assigned attention head $h$, where $c$ indexes the class and $h \in \{1, ..., N\}$ indexes the head. For example, with $\mathcal{C}=200$ classes and $N=10$ heads, we have 2,000 prototypes total: the first 10 prototypes belong to class 1 and are compared with heads 1 through 10 respectively, the next 10 prototypes belong to class 2, and so on. This architectural constraint ensures each prototype specializes in features captured by its assigned attention head, preventing multiple prototypes from converging to similar representations. We explore $D_1 \in \{32, 16, 8\}$ across different backbones.

\subsection{Classification}

The similarity scores are combined to produce class logits through a fully-connected layer. For each class $c$, the logit is computed as a weighted sum of similarity scores to the $N$ prototypes assigned to that class:

\begin{equation}
\text{logit}_c(\mathbf{x}) = \sum_{h=1}^N w_{c,h} \cdot s_{h,j_c^h}(\mathbf{x})
\end{equation}

where $w_{c,h}$ are learnable weights and $j_c^h$ denotes the prototype index for class $c$ and head $h$. The final prediction is obtained via softmax over all class logits. This formulation ensures each classification decision aggregates evidence from diverse attention heads, with each head contributing through its specialized prototype.

\subsection{Contrastive Regularization Details}

While the one-to-one architectural constraint provides a strong inductive bias for diversity, it alone does not guarantee that prototypes will remain distinct during training. Without additional regularization, prototypes assigned to different attention heads may still converge to similar representations if those features are highly discriminative for classification. This is particularly problematic when certain visual patterns (e.g., wings in bird classification) are more salient than others, causing the optimization to favor redundant solutions despite the architectural separation.

The contrastive loss (Eq.~3 in the main paper) operates on the similarity scores, where $s_{h,i}$ is the pooled similarity between attention head $h$ and prototype $i$, $\mathbf{P}_{\bar{h}}$ denotes prototypes not assigned to head $h$, and $m$ is a margin parameter. This maximizes similarity between assigned prototype-head pairs while minimizing similarity to unassigned prototypes. In practice, we sample a subset of negative prototypes per batch to reduce computational cost.

\section{Detailed Evaluation Metrics}

\subsection{Prototype-Part Assignment}

We compute an overlap score between each prototype's activation and part mask:

\begin{equation}
\text{overlap}(\mathbf{p}_j, \text{part}_k) = \frac{|A_j \cap M_k|}{|M_k|}
\end{equation}

where $A_j$ is the thresholded activation region of prototype $j$ and $M_k$ is the binary mask of part $k$. We normalize by part size to ensure small parts (e.g., beak) are not underweighted. A prototype is assigned to a part when $\text{overlap}(\mathbf{p}_j, \text{part}_k) > \tau$.

\subsection{Coverage Score}

Coverage measures whether prototypes capture semantically meaningful parts using precision and recall:

\begin{equation}
\text{Precision} = \frac{|\{\mathbf{p}_j : \mathbf{p}_j \text{ assigned to } \geq 1 \text{ part}\}|}{|\mathcal{P}|}
\end{equation}

\begin{equation}
\text{Recall} = \frac{|\{\text{part}_k : \text{part}_k \text{ captured by } \geq 1 \text{ prototype}\}|}{K}
\end{equation}

\begin{equation}
\text{Coverage} = \text{F1} = \frac{2 \cdot \text{Precision} \cdot \text{Recall}}{\text{Precision} + \text{Recall}}
\end{equation}

where $|\mathcal{P}|$ is the total number of prototypes and $K$ is the number of semantic parts.

\subsection{Diversity Score}

Diversity measures prototype diversity by combining spatial redundancy and part coverage balance. Redundancy quantifies spatial overlap between prototypes assigned to the same part:

\begin{equation}
\text{Redundancy} = \frac{1}{|R|} \sum_{r \in R} \frac{1}{C_r} \sum_{(i,j) \in \text{pairs}(P_r)} \text{IoU}(A_i, A_j)
\end{equation}

where $R$ is the set of parts, $P_r$ is the set of prototypes assigned to part $r$, and $C_r$ is the number of prototype pairs for part $r$. Note that redundancy scores depend on the quality of prototype-part assignments and may be sensitive to the threshold $\tau$. Imbalance measures uneven distribution of prototypes across parts using the Gini coefficient:

\begin{equation}
\text{Imbalance} = \frac{\sum_{i=1}^{K}\sum_{j=1}^{K}|c_i - c_j|}{2K^2\mu}
\end{equation}

where $c_k$ is the number of prototypes assigned to part $k$ and $\mu$ is the mean number of prototypes per part. These are combined into the Diversity score:

\begin{equation}
\text{Diversity} = 1 - \frac{\text{Redundancy} + \text{Imbalance}}{2}
\end{equation}

High Coverage indicates prototypes capture meaningful semantic features; high Diversity indicates prototypes are spatially distinct with balanced part coverage.

\section{FaceForensics++ Results}

DAPL generalizes beyond fine-grained classification to facial manipulation detection on FaceForensics++~\cite{ff++} (frames uniformly sampled from videos: 296,238 training, 57,900 test across 4 manipulation types). Foreground masks are generated using MediaPipe landmarks~\cite{mediapipe} refined with SAM~\cite{SAM}. We use 50 prototypes (10/class) with assignment threshold $\tau=0.5$.

\begin{table}[t]
\centering
\caption{Performance on FaceForensics++ (DenseNet161).}
\label{tab:main_ff}
\begin{tabular}{lccc}
\toprule
\textbf{Method} & \textbf{Accuracy} & \textbf{Coverage} & \textbf{Diversity} \\
& (\%) & $\uparrow$ & $\uparrow$ \\
\midrule
ProtoPNet & 93.24 & 0.288 & 0.312 \\
\midrule
DAPL ($D_1$=32) & 93.19 & \textbf{0.654} & 0.294 \\
DAPL ($D_1$=16) & 93.25 & 0.567 & \textbf{0.378} \\
DAPL ($D_1$=8) & \textbf{93.75} & 0.615 & 0.315 \\
\midrule
DAPL ($D_1$=16) + FG & 93.66 & 0.478 & 0.318 \\
\bottomrule
\end{tabular}
\end{table}

DAPL demonstrates consistent effectiveness on facial manipulation detection (Table~\ref{tab:main_ff}). With DenseNet161 ($D_1=16$), DAPL achieves \textbf{93.25\% accuracy}. Foreground-aware training maintains strong accuracy at \textbf{93.66\%} (+0.42\% over ProtoPNet) while achieving improved interpretability: Coverage improves by \textbf{+66\%} (0.288 $\rightarrow$ 0.478) and Diversity by \textbf{+42\%} (0.312 $\rightarrow$ 0.442).

ProtoPNet struggles on this task: 8 out of 10 prototypes activate on background regions, with only 2 capturing facial features (both on cheeks). This severe background bias prevents the model from learning manipulation-relevant features. DAPL addresses this limitation, with prototypes covering 6 distinct facial regions (cheeks, nose, eyes, mouth, eyebrows). Notably, DAPL ($D_1=32$) achieves the highest Coverage (0.654), demonstrating that different attention dimensions offer different interpretability-accuracy trade-offs.

\begin{figure}[t]
\centering
\includegraphics[width=\linewidth]{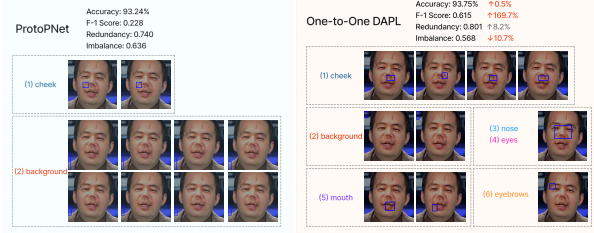}
\caption{\textbf{Prototype activation comparison on FaceForensics++.} ProtoPNet (left) shows redundancy with 2 cheek and 8 background prototypes. DAPL (right) achieves diverse coverage across 6 facial regions (cheek, nose, eyes, mouth, eyebrows). Coverage (F-1 Score) improves by 169.7\%, demonstrating substantially better semantic alignment.}
\label{fig:ff_comparison}
\end{figure}

\section{Detailed Ablation Studies}

We provide extended analysis of the ablation studies presented in the main paper (Figure 3).

\textbf{Contrastive Regularization Coefficient.} The coefficient $\lambda_c$ controls the strength of diversity enforcement between attention heads (Figure~\ref{fig:ablation_studies}, left). At very low values ($\lambda_c=0.01$), the architectural constraint alone provides insufficient diversity, resulting in lower Coverage (0.523). As $\lambda_c$ increases to 0.5, Coverage peaks at 0.600, indicating optimal prototype specialization. However, excessively high values ($\lambda_c \geq 1.0$) over-constrain the model, forcing prototypes into overly distinct representations that may not align with natural visual patterns, reducing Coverage to 0.434-0.579. The optimal $\lambda_c=0.5$ balances diversity enforcement with classification performance, achieving 80.98\% accuracy.

\textbf{Foreground Loss Coefficient.} With background threshold fixed at $\tau_{bg}=0.3$, we vary $\lambda_{fg} \in [0.5, 25.0]$ (Figure~\ref{fig:ablation_studies}, middle). Accuracy steadily improves from 81.17\% to a peak of 81.95\% at $\lambda_{fg}=15.0$, demonstrating that foreground guidance helps the model focus on discriminative bird features. Coverage follows a similar trend, increasing from 0.446 to 0.596 at $\lambda_{fg}=10.0$, then plateauing. Beyond $\lambda_{fg}=15.0$, both metrics slightly decline as excessive foreground constraints over-specialize prototypes. The results reveal that moderate foreground guidance ($\lambda_{fg}=10.0$-$15.0$) provides the best interpretability-accuracy balance for fine-grained classification.

\textbf{Background Threshold.} With $\lambda_{fg}=20$ fixed, we examine how background tolerance threshold $\tau_{bg}$ affects performance (Figure~\ref{fig:ablation_studies}, right). Stricter thresholds (lower $\tau_{bg}$) force prototypes to activate primarily on foreground regions. Accuracy improves monotonically from 80.79\% ($\tau_{bg}=0.01$) to 82.02\% ($\tau_{bg}=0.25$), while Coverage increases from 0.552 to 0.584. This trend indicates that allowing minimal background activation ($\tau_{bg}=0.20$-$0.25$) is beneficial, as it provides flexibility for prototypes to capture contextual information at object boundaries while maintaining semantic focus. Overly strict thresholds ($\tau_{bg} < 0.10$) may exclude valid boundary features.

\section{Foreground Mask Generation}

\textbf{CUB-200-2011:} We use the provided binary segmentation masks that distinguish bird pixels from background.

\textbf{FaceForensics++:} We generate foreground masks in two steps: (1) Detect 468 facial landmarks using MediaPipe~\cite{mediapipe}. (2) Create an initial mask by computing the convex hull of landmark points. (3) Refine mask boundaries using SAM~\cite{SAM} with the convex hull as a prompt, producing precise foreground segmentation that captures facial contours while excluding background regions.

\subsection{Foreground-Aware Projection and Pruning}

\textbf{Projection Phase:} During prototype projection (replacing prototype vectors with nearest training patches), we only consider patches where the foreground overlap ratio exceeds threshold $\tau_{\text{FG}}$:

\begin{equation}
\text{FG\_Ratio}(\mathbf{z}) = \frac{\sum_{(i,k) \in \text{patch}(\mathbf{z})} M(i,k)}{|\text{patch}(\mathbf{z})|} > \tau_{\text{FG}}
\end{equation}

We use $\tau_{\text{FG}} = 0.5$ for both datasets, ensuring prototypes are grounded in patches with majority foreground content.

\textbf{Pruning Phase:} After projection, we compute each prototype's average bias score across the validation set. Prototypes with bias scores below threshold $\tau_{\text{bias}} = 0.3$ are removed, as they consistently activate on background regions rather than semantically meaningful object parts.

\section{Domain-Specific Hyperparameter Discussion}

Optimal hyperparameters vary by domain, reflecting task-specific requirements. Fine-grained bird classification benefits from moderate attention dimension ($D_1=16$) that balances detailed features with contextual information, while facial manipulation detection requires higher specialization ($D_1=8$) to capture subtle artifacts. Similarly, foreground guidance requirements differ: CUB-200-2011 achieves best results with $\lambda_{fg}=10.0$-$15.0$, while FaceForensics++ benefits from stronger constraints ($\lambda_{fg}=20.0$) to avoid background bias.

\end{document}